# A utility belt for an agricultural robot: reflection-in-action for applied design research

**Natalie Friedman**
Cornell Tech
nvf4@cornell.edu

**Asmita Mehta**
IIT Dehli
asmita8mehta@gmail.com

**Kari Love**
New York University
klove@nyu.edu

**Alexandra Bremers**
Cornell Tech
awb277@cornell.edu

**Awsaf Ahmed**
Brooklyn Tech
awsafa58@gmail.com

**Wendy Ju**
Cornell Tech
wendyju@cornell.edu

**ABSTRACT**
Clothing for robots can help expand a robot's functionality and also clarify the robot's purpose to bystanders. In studying how to design clothing for robots, we can shed light on the functional role of aesthetics in interactive system design. We present a case study of designing a utility belt for an agricultural robot. We use reflection-in-action to consider the ways that observation, in situ making, and documentation serve to illuminate how pragmatic, aesthetic, and intellectual inquiry are layered in this applied design research project. Themes explored in this pictorial include 1) contextual discovery of materials, tools, and practices, 2) design space exploration of materials in context, 3) improvising spaces for making, and 4) social processes in design. These themes emerged from the qualitative coding of 25 reflection-in-action videos from the researcher. We conclude with feedback on the utility belt prototypes for an agriculture robot and our learnings about context, materials, and people needed to design successful novel clothing forms for robots.

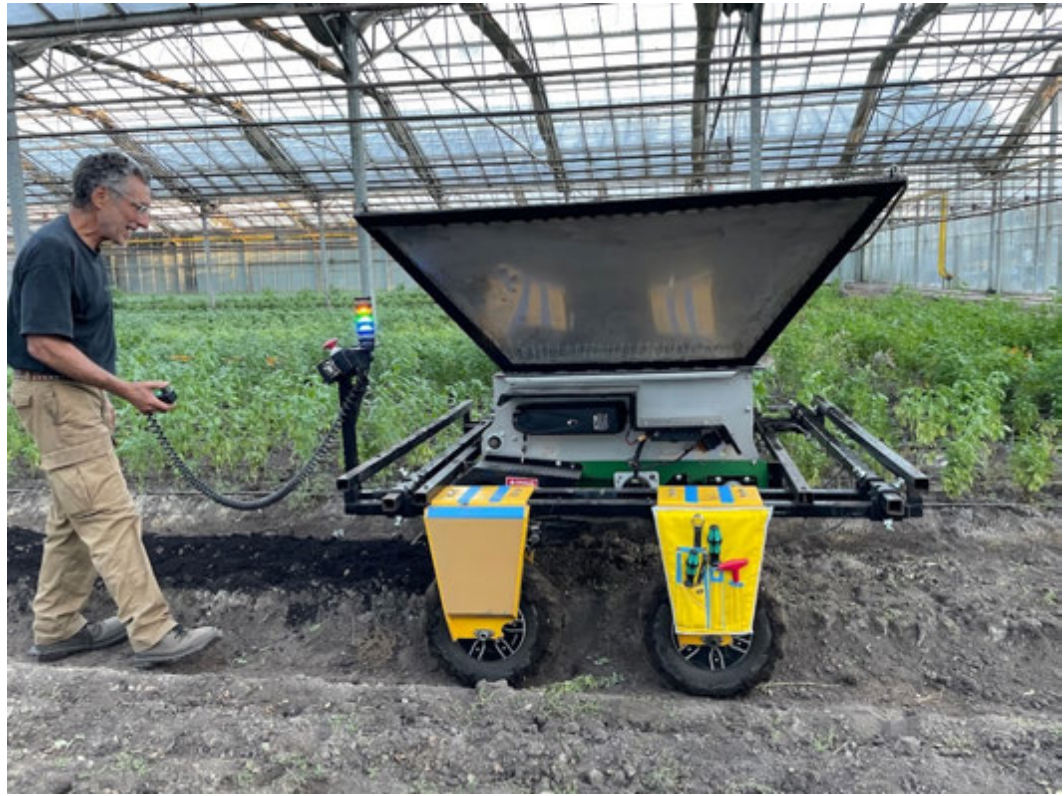

Fig. 1. Larry Jacobs, owner of Jacob's Farm, operating the farm robot called "Amiga" which is wearing a prototype of our utility belt.

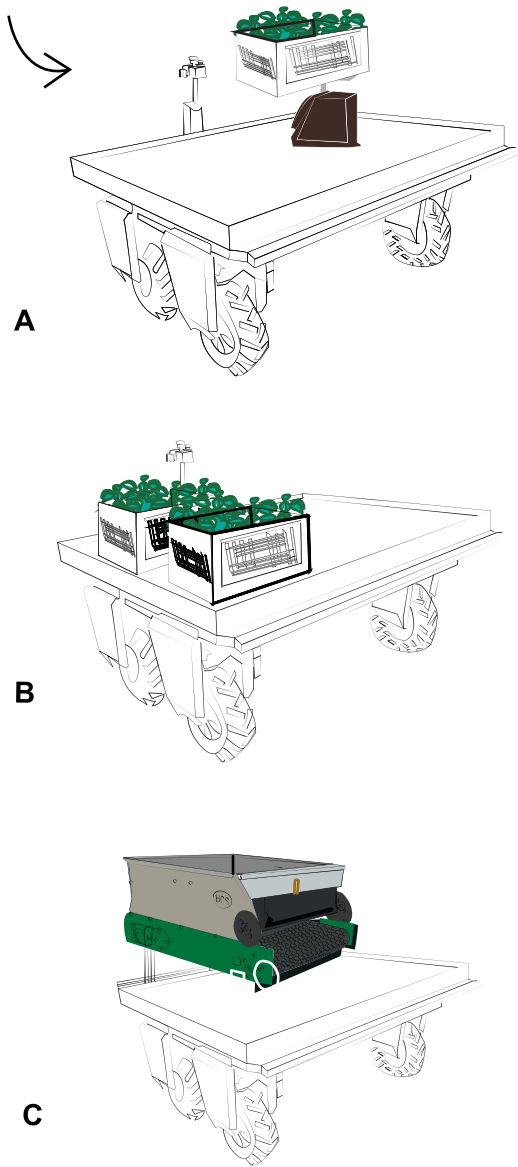

*The Amiga and it's various capabilities*

# 1 INTRODUCTION

How can we bring design research into the field? In situ design can benefit research through design by immersing the researcher in the context of design, helping her to build relationships and understandings with potential users, and enabling rapid insight-prototype-test cycles. In this pictorial, we illustrate the process of research-through-design through a case study of our development of a utility belt for an agricultural robot.

## 1.1 Design Objective

The robot we were designing for is called ``the Amiga,'' a battery-powered, small electric tractor robot. (See Figure 1.) It was built by the robotics start-up Farm ng as an open-source platform to be adapted by farmers to assist in a variety of farm tasks, like compost spreading (Figure 2C), pulling other machinery (Figure 2D), or carrying vegetable totes (Figure 2B). The decision to focus on developing robot clothing occurred at the project's outset, as it was my (the first author's) Ph.D. research topic.

The utility belt was inspired by the observation that people were placing their tools atop the robot and that "a friend" could help carry these things for workers. A utility belt could be taken on and off and help make the robot more versatile.

## 1.2 Design Site

The focus of this pictorial is our reflections on the challenges and opportunities presented by designing the robot's utility belt in situ. The project's site was Jacobs farm, an organic farm in Watsonville, California. The farm owners were open to pilot testing the Amiga because they were interested in using clean energy, preventing pests, and making their farms more efficient. We worked with and among farm workers and roboticists on this project. Culturally, this project fits in with the other work going on; the open-source roboticists and farmers were surprisingly alike in their comfort with in situ design.

In addition to working at Jacob's Farm, we also worked at the offices of Farm ng. We got help from people at a Maker Space in Felton, California, called Crooked Beauty, and a fabric co-op in Santa Cruz, California, called the Fabrica.

**Fig. 2.** The robot can carry (A) a scale and a full box of basil, (B) two crates of basil, (C) a compost spreader, (D) pull a crate, (E) connect with a weeder.

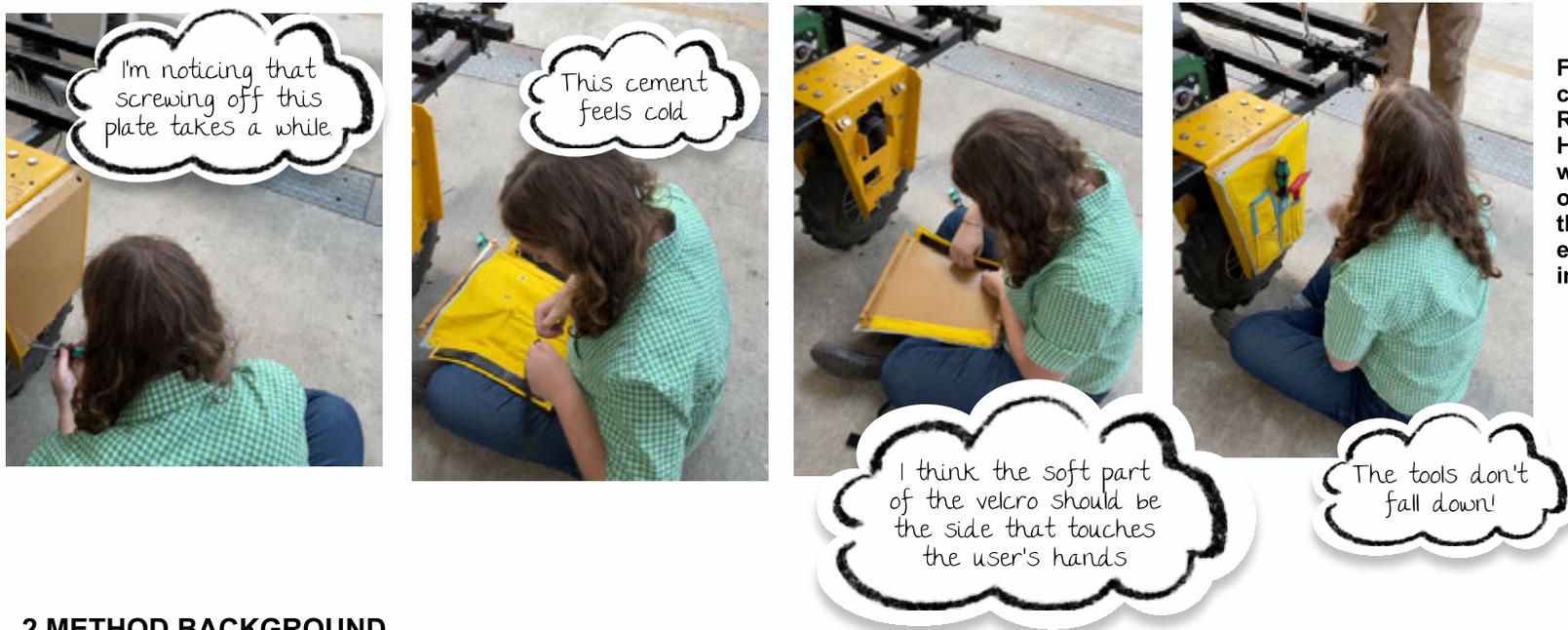

Fig. 3. Reflection-in-action can be a rich method for Research through Design. Here, the researcher reflects while putting the utility belt on the robot, contemplating the material and how her environment supports or impedes her efforts.

## 2 METHOD BACKGROUND

### 2.1 Reflection in Action

Donald Schoen described reflection-in-action as a way for designers to develop theory from practice to convert tacit knowing-in-action to more explicit understandings, which can then be surfaced, criticized, restructured, and embodied in further action [16]. Reflection-in-action, then, is a Research through Design method and requires the designer/researcher to be mindful of the research theories and techniques even as they are making in-the-moment judgments and performances. While Zimmerman et al.'s articulation of RtD is very artifact-centered [19], Schoen foundationally describes the design as "a reflective conversation with the situation" [16] and, as such, we argue that performing and documenting reflection-in-action for design as it occurs in situ should be a standard mode of RtD in interactive system design.

### 2.3 Designing in situ

The practice of contextual inquiry [17,1] and in situ evaluations are well-established in CHI and DIS [4, 10, 5], but in situ design is less common in the literature, even though it is often common practice in industry. By "in situ design," we mean that the designers are performing the steps of ideation, prototyping, and early-stage testing at a site where the products of the design activity would be used; this is different than merely observing the practices and context at the site, or trying out and evaluating prototypes at a site [15]. This approach is more similar to ethnomethodological research, wherein end-users perform design [7, 3] or where autobiographical design and research occur [2].

On the other hand, there are fields of HCI where in situ design is common. For example, in the field of Information and Communications Technologies for Development (ICT4D), Dearden and Rivzi noted that "to be sustainable, intervention needs to be compatible with the constraints [imposed by the site]. The sustained, regular presence of the research in the field site is critical to maintaining trust and sensitivity to these complex factors" [6]. Similarly, Halloran et al. have noted the importance of co-designing ubiquitous computing-supported tours in situ, over time, to understand the longitudinal issues of maintenance and adaptation of technologies, as well as the socio-technical effects of those technologies on both users and curators [11].

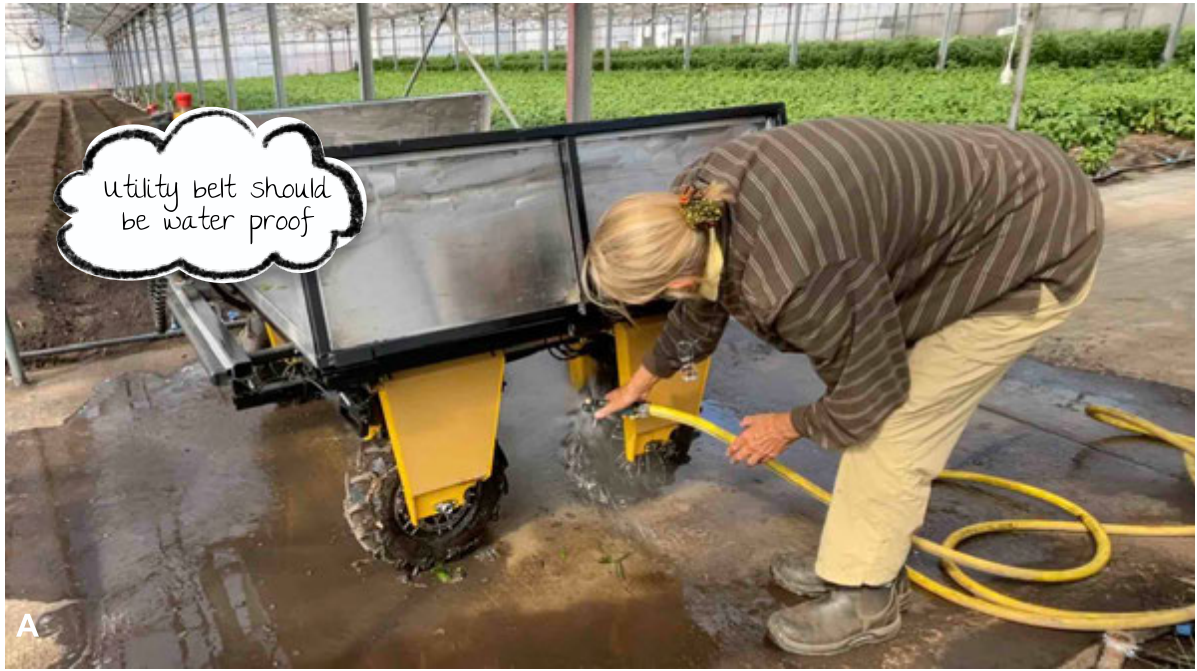
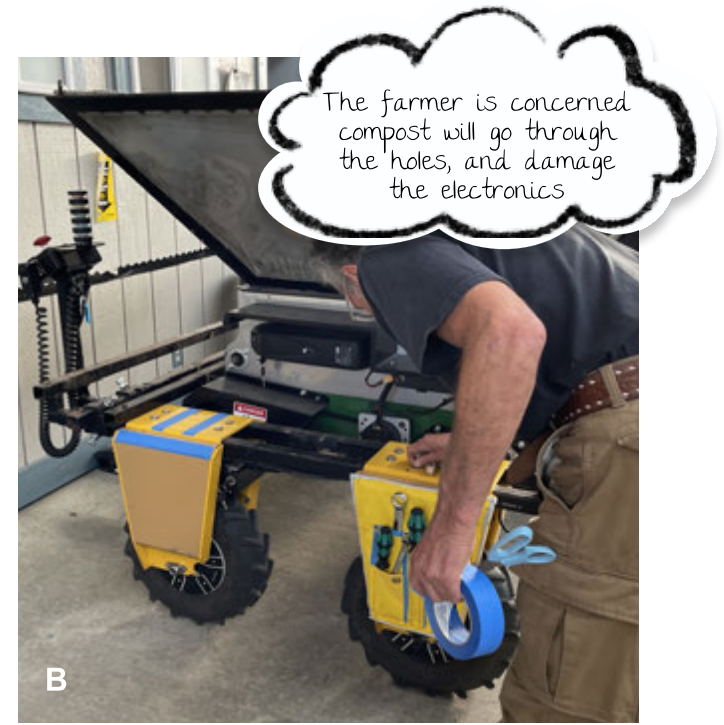

**Figure 4.** (A) Jacob's Farm co-owner hosing down the wheels on the Amiga, posing an issue of water getting on the electronics.
(B) Jacob's Farm co-owner using tape to protect the Amiga from compost

## 3 CONTEXTUAL DISCOVERY

To design clothing for the Amiga, we spent time understanding the farm context; specifically, we explored everyday practices (Section 3.1), existing farm materials (Section 3.2), tools (Section 3.3), and lastly, the current culture on the farm (Section 3.4).

### 3.1 Understanding farmer practices with Amiga

I extensively followed the robot in use. I watched the robot follow the farmers while they cut basil. They walked up and down the greenhouse rows, and adapted relatively quickly to incorporating the robot into their work. When the automatic following didn't work perfectly, we had them teleoperate the robot.

While it was helpful to see the everyday practices, outlier moments, like washing (Figure 4A), brought us the richest design insights.

During one contextual inquiry session, I watched a farmer owner hose down the robot. I knew this robot version could not handle all that water yet; the electronics had not been waterproofed. But the farmer hosed off the robot with pride, just like she did all her other equipment. I watched in horror, paralyzed. However, later, another member of the research team was excited by the moment: "I love that farmer who is washing it! Users tell us how to design by making these kinds of mistakes!" Later versions of the robot would have watertight plates.

When the farm owner mounted the compost spreader on the robot, he covered the screw holes with tape so that the compost would not get into the electronics. I love to be surprised by how the user "hacks" the robot with materials available to them. I learned that the farmers understood that the robot could be harmed by the compost. By being there, I witnessed these precautions and used them for design insights.

## 3.2 Exploring existing farm materials

Because the farm robot is exposed to water and sunlight, we needed to consider what materials the robot clothing should be made of. We studied the materials used in clothing and equipment for farmers to understand issues and pre-existing farm solutions better. We visited Tractor Supply Co, a farm goods store in Watsonville, California, which sold everything from feed and fencing to farm clothes. We documented the sturdiness of existing utility belts (Figure 5D), the feeling of touching a glove (Figure 5A), and used jumpsuits (Figure 5B).

We also documented the colors and textures used in existing goods as a reference for designing the palette for the robot's belt. We felt, for example, that we could use materials and colors like those used in the farm workers' own clothes to establish a sense of team membership among the farmers and robots.

After our trip to Tractor Supply Co, often, the discussions about the materials revealed underlying concerns that each stakeholder paid attention to. For example, one engineer talked about how he wished a utility belt would be designed with transparent material to allow the user to see the size of the particular wrench and to grab the wrench safely. However, the other robot designer who was onsite was concerned by this suggestion because clear plastic gets damaged easily and could make the robot look less slick. This pointed out to us the issues of utility, transparency, and efficiency and how they might be in tension with durability and aesthetic considerations.

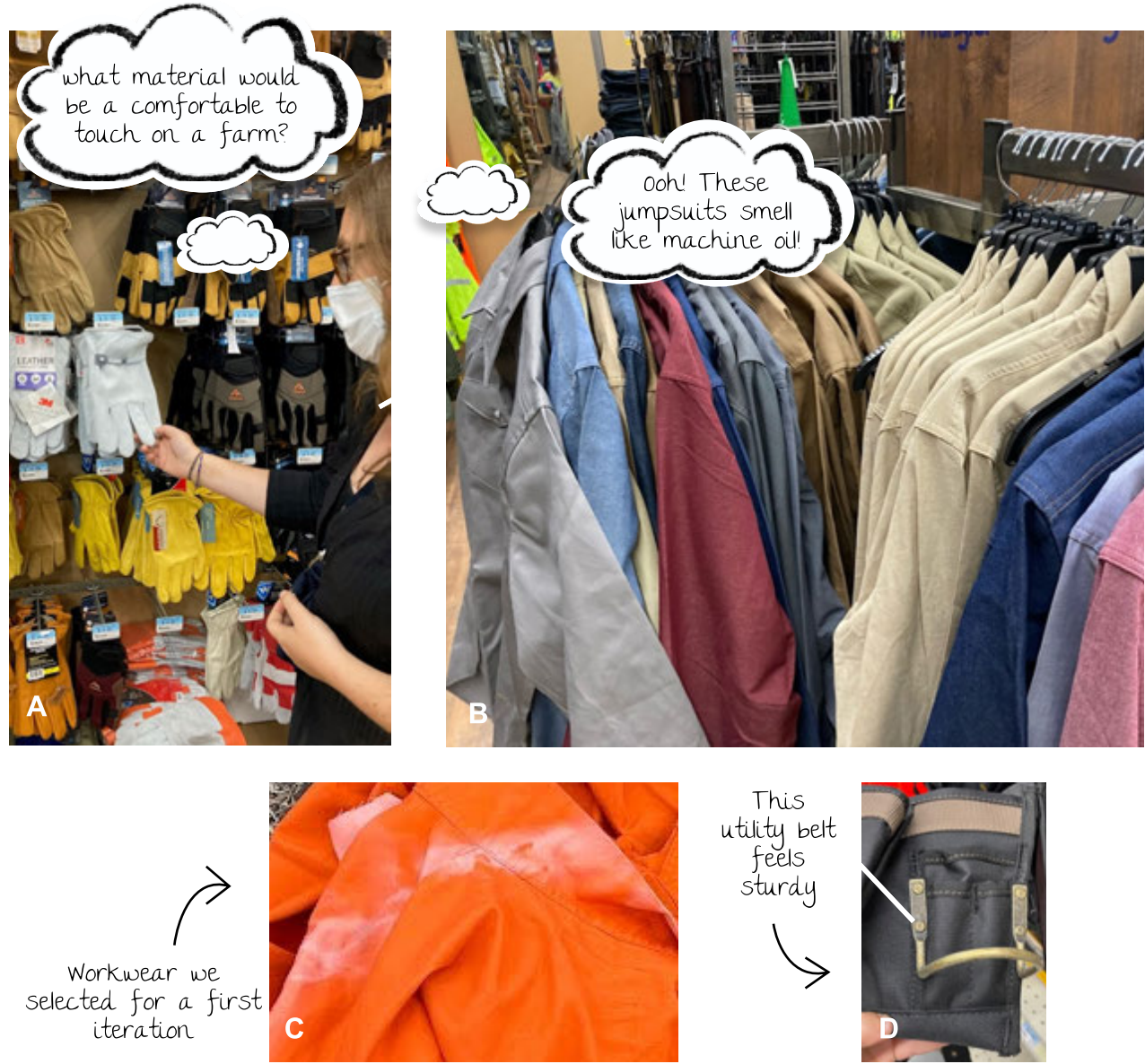

**Fig. 5.** A trip to the Tractor Supply store in which we documented: (A) Variety of gloves, (B) Used workwear, (C) Workwear selected for first iteration, (D) Utility belt

## 3.3 Exploring existing farm tools

To establish which tools would go into our robot's utility belt, I took an inventory of current tools (sockets, screwdrivers, etc.) and tool belts used at Farm ng and Jacob's Farm (Figure 6A-6D). Farmers carried items like scissors, voltage testers, and mobile phones. I measured both the tools at Farm ng and Jacob's Farm in preparation for pockets on a tool belt. Once the robot came into the picture, however, there were additional tools that the farmers might need to carry. To change the width and length of the robot for different tasks, the farmers need screwdrivers and wrenches. We suggest that the robot should carry its own tools.

## 3.4 Exploring farm culture

At the farm, I learned about the culture of work and the culture of the people who worked there. For example, the farm manager organized decorations in the farm corridor for Día de los Muertos, with photos of family who had died of Covid-19 that year (Figure 7A). As a designer, it was helpful to understand the culture in which I was creating in. I could consider the colors for this celebration and intentionally not use them, as those would be reserved for their celebration and not my design.

We can learn from the way the farmers were naturally accessorizing the robot. In the image to the right, the farmers hang fabric speaker cases on the robot's chassis and play music while the robot follows them (Figure 7B). When a farmer accessorized the robot with a speaker, we could learn which tasks the physicality of the speaker would not get in the way, and the music could also be heard.

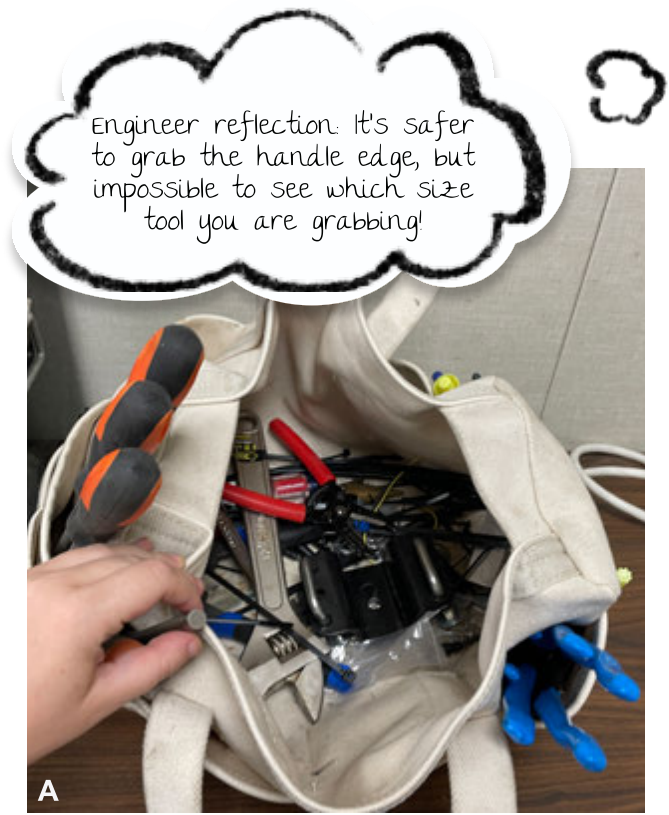

Fig. 6. (A) Inside a tool bag, (B) Voltage tester held by farm engineer, (C) Plant scissors held by farmer, (D) Scissors on a mobile cart

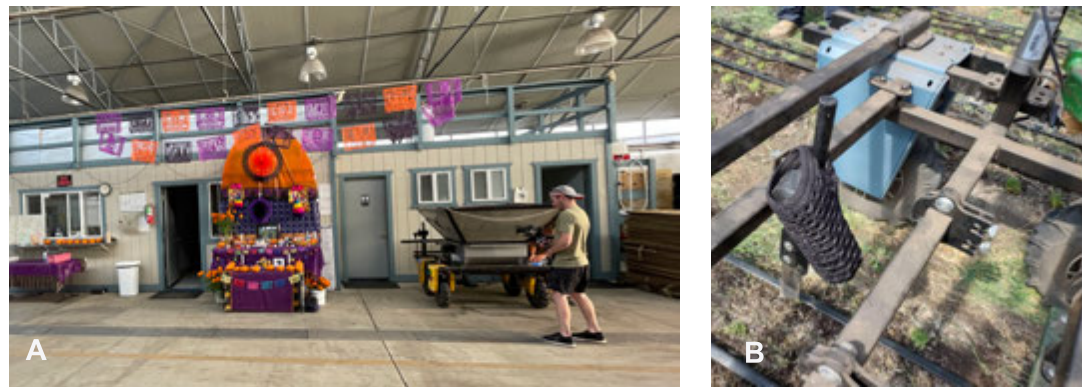

Fig. 7. (A) Decorations for Día de los Muertos at Jacob's Farm, (B) Speaker hanging on Amiga

## 4 DESIGN

After using ethnographic methods to observe the farmers working with the robot and inspect the tools on the farm, I was ready to design.

### 4.1 Designing functional clothes for robots

Designing a utility belt was part of a larger project on designing clothing for robots. Because of a utility belt's removability, we consider it clothes or accessories. Some of the lessons going into this project are discussed below.

Let us first consider what robot clothing entails. Robot clothes should be motivated by what robots and their users need. While clothes for robots should not mimic clothes for people, we can draw from the functions of clothing for people to design clothing for robots. For example, clothes for people can (1) **signal** where a person belongs, (2) can **protect** their skin from getting scratched, and (3) can help them **adapt** to whatever setting might come next, ideas outlined in [8]. Respectively, for an agriculture robot, coverings that are a similar color and material can help the robot socially **signal** that they are part of the farm team, can **protect** it from water or soil, and help the robot **adapt** to any given task, as shown in Figure 8A.

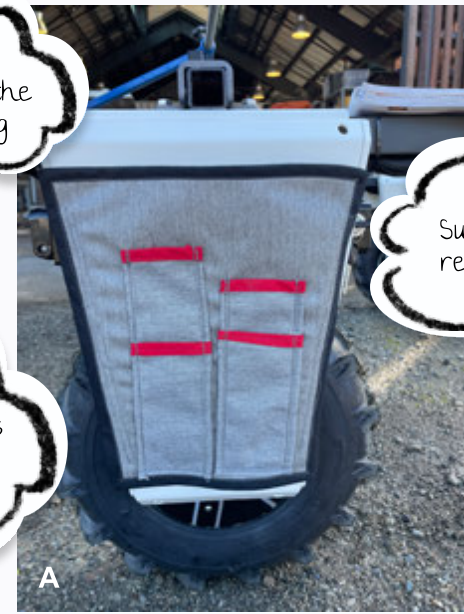

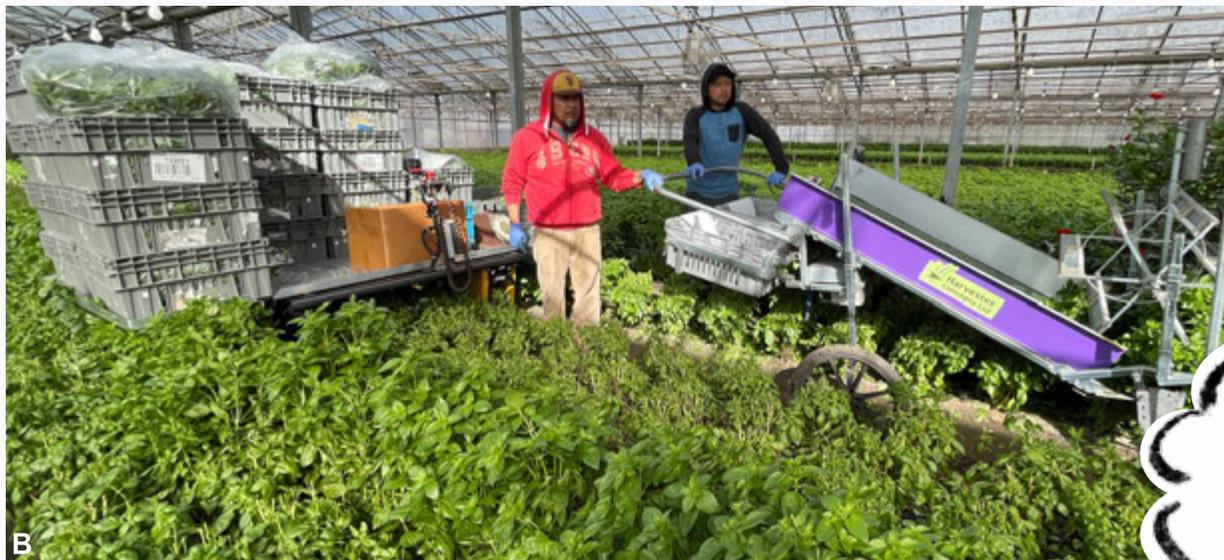

Fig. 8.
(A) Final utility belt provides social, signaling adaptability, and protection,
(B) Robot carrying totes of basil

## 4.2 Materials in situ

In exploring materials for the farm belt, I wanted to think about how those materials interact with the environment that the belt would be used. In this exploration, I made a point of using my senses, thinking bout how the material looked, felt, and even smelled in use and in context. I wanted the robot to not only functionally fit in but also to fit in aesthetically. I, therefore, set out to understand the aesthetic of the fabric in the field.

*Material Selection.* Fashion writer Elizabeth Hawes describes, "To dress fashionably is both to stand out and to merge with the crowd" [12]. Inspired by this, we selected materials that covered the design space between fitting in (like workwear, pictured in Figure 9B) and standing out (like neon green fabric, pictured in Figure 9C), both functionally and aesthetically.

*Learnings from materials.* Ripstop (Figure 9C), indicated by its name, is resistant to ripping. Additionally, many other features could help a robot in an agriculture setting; it can be waterproof, water and fire-resistant, and have low or no porosity, meaning water or air won't penetrate the fabric. As suggested by a costumer, the neon green gave both visibility to the robot and pointed out that the robot is not part of the natural environment. This clashing may have failed to help the robot belong. Similarly, this was true about the metallic fabric (Figure 9A). Therefore, we decided to go with the orange workwear over the neon green ripstop or metallic. Comparing these materials in situ helped us understand the benefits of some fabrics over others and the reasoning for why these materials get sold in a farm goods store.

*Methodology insight:* We could see the juxtaposition of colors and textures on the dirt and grass. We could learn, do the grass and dirt get stuck in the fabric. Does thread fall off the fabric and fall onto the grass?

*Methodology insight:* Designing in situ allowed us to use our senses. We could feel while donning and doffing the material onto the robot, a temperature change when the material was placed on the robot, and the dirt on our knees while kneeling. We and the material were having a conversation, as suggested by Schon [16].

I see . . . . . textures of fabric, sky, trees, grass

I smell . . . . . scents of fresh air and grass

I hear . . . . . . . . sounds of wind

Fig. 9. We laid fabric on the field, including:
(A) Metallic cloth,
(B) Traditional orange workwear,
(C) Bright green ripstop,
(D) A variety of other fabric

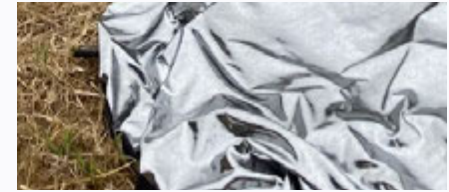
A

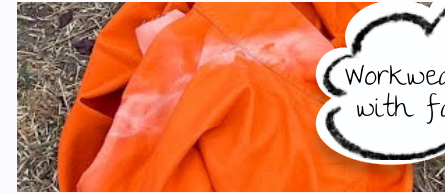
B

*Workwear blends in with farmer wear*

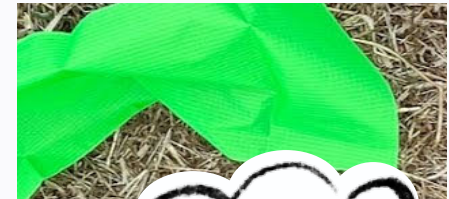
C

*These fabric colors do not appear in nature*

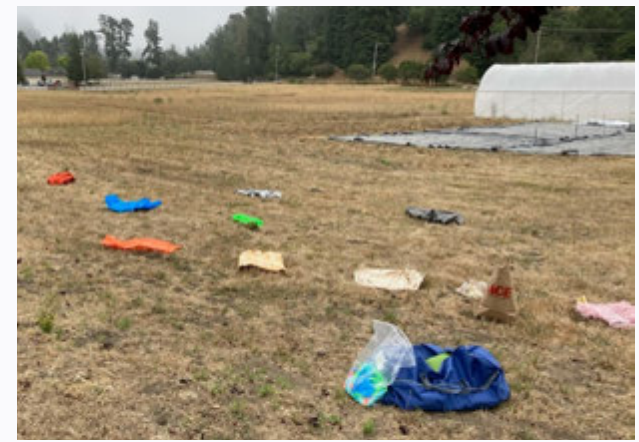
D

### 4.3 Making and improvising

We use low-fidelity prototyping, which is affordable (often involves cardboard and office supplies) and allows for quick making [18]. I iterated on the utility belt with materials I had discovered during my in situ explorations in Section 4.2. For example, materials I explored included workwear (Figure 11B), for team membership and ripstop (Figure 11C), for durability.

Because I was designing in the field, I did not have access to the toolkit of prototyping supplies in a design studio. I needed to do more to improvise; for example, I discovered that whiteboard magnets in the Farm ng office (Figure 11C) could be used to help hold the fabric in place. Other serendipitous prototyping materials included plastic forks, which I used as hooks (Figure 11A), and binder clips (Figure 11B). The design constraints of a limited selection helped me be creative in achieving different forms of the utility belt quickly.

I customized the pockets (Figure 10) to fit the size of the tools for changing the width and length of the robot, which, in turn, could make the robot more adaptable for the farmers. Unfortunately, I did not always have a seamstress friend to help me make a utility belt. Although I am not a professional seamstress, I could prototype quickly in the moment, which was still extremely beneficial.

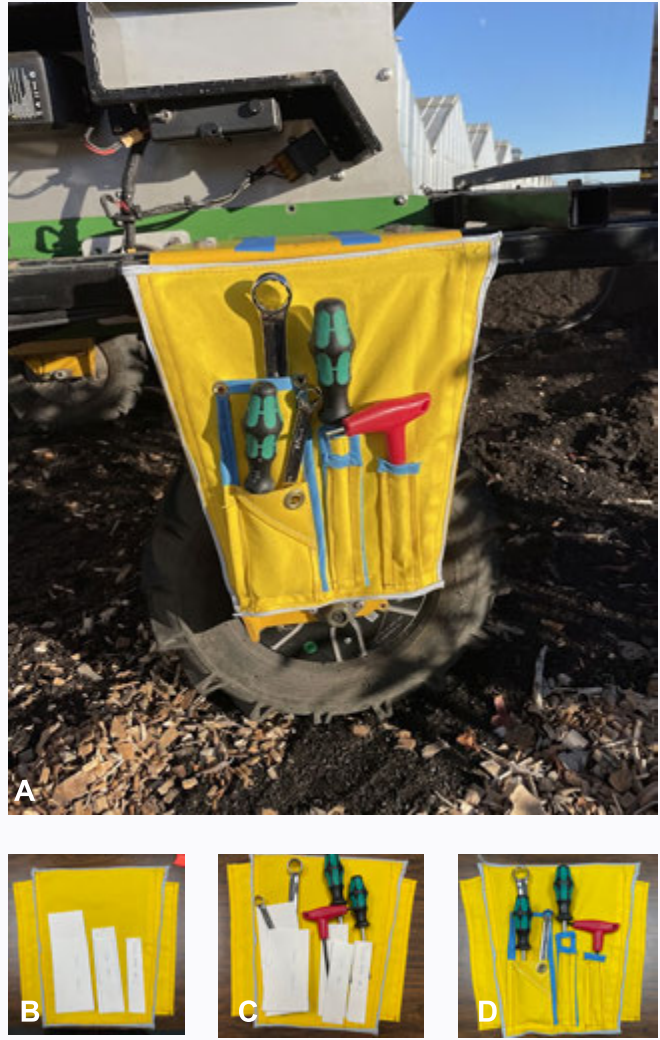

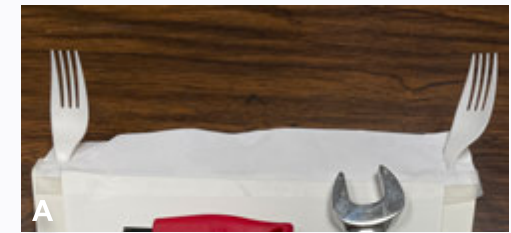

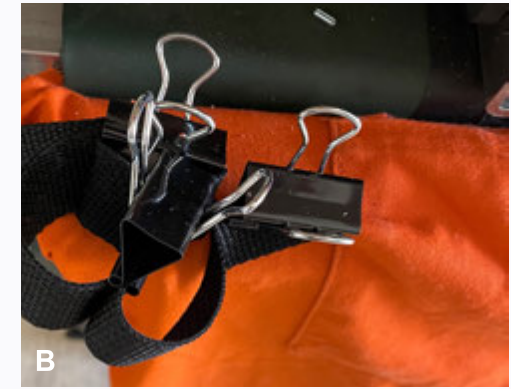

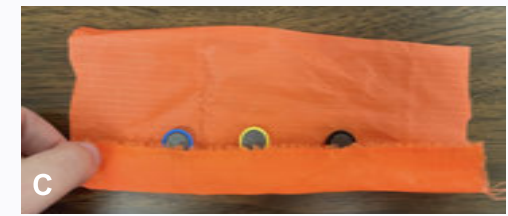

Fig. 10. Design process of prototype 2:
(A) Robot wearing the utility belt on the farm
(B) Tool belt preparation with paper cut outs of pockets
(C) Tool belt paper cut outs of pockets with tools
(D) Tool belt with tools laid down on a conference table

Fig. 11. Using office supplies to prototype at Farm ng:
(A) Plastic forks being used as hooks,
(B) Binder clips attaching the workwear to the robot,
(C) Magnets being wrapped in fabric

# 5 DISCUSSION: REFLECTIONS FROM ACTION

## 5.1 Making space for inquiry and making

During the observations, I often found myself in the way. I needed to interrupt engineers during their robot production to ask questions about the purpose of each tool. My explorations were sometimes a hindrance. One time, the robot engineer looked everywhere for a tool I had borrowed. I had the tool in my workspace to prototype pockets for the utility belt. While my removal of this tool was an interference, it also taught me about the priority of the tool for its original purpose, outside of its role as a form or prop for my belt design activities. To do design in situ, I had to make maker spaces out of non-maker spaces. For example, in the images in Figure 12, one can see that I turned a robotics company's conference room into a space for fabricating textile products. I turned the farm's cement corridor into a dressing room, where I knelt and dressed the robot plates. I made the greenhouse rows into a place of research in which I documented design discussions.

Working in situ, I had to learn to make space for my research and inquiry. As I was learning what each tool was called and how it was used, I felt like I was learning how to use tools from a parent, in a way. It felt like I was the wrong gender, in the wrong place and time. This is a natural consequence of working in situ; these spaces were not originally designed for the work I was doing there. Over time, I became more comfortable taking up space, asking these questions, and designing in front of my non-design colleagues. I also got to have discussions about what it was I was doing, and that also helped my own sense of process and reflection.

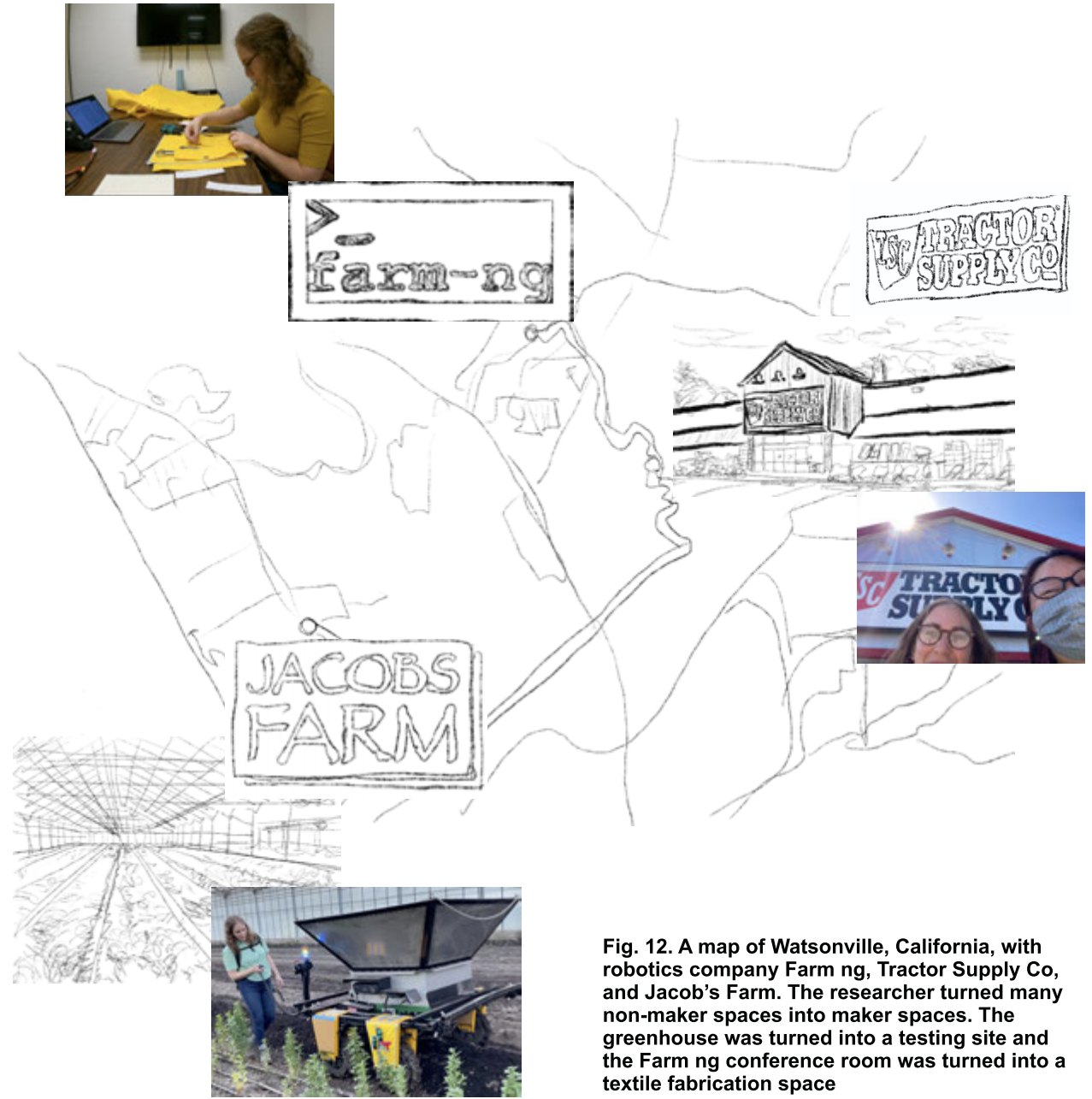

Fig. 12. A map of Watsonville, California, with robotics company Farm ng, Tractor Supply Co, and Jacob's Farm. The researcher turned many non-maker spaces into maker spaces. The greenhouse was turned into a testing site and the Farm ng conference room was turned into a textile fabrication space

## 5.2 Social process of design

Other major reflections which occurred to us throughout the design process was how social the design process was. I reviewed the design decisions I was making in the moment with a larger team that was not on-site. For example, when I showed one of the prototypes (Figure 13C) to a team member who had a costuming and soft robotics background, she questioned - why is there a belt across the two forks? Why this material? It seemed like I was basing the utility belt on the form of a human's belt. But for a robot, it was another hanging extension that could get caught on grass or other farm equipment. She also pointed out that the canvas fabric could rot over time, while the plushness looked like it could get dirty easily. We discussed these points abstractly before I went on site; I should design for the robot's needs and not from the past perception of how people wear clothes. Putting this into practice was surprisingly challenging, and it points out that reflection is not always self-reflection but a social reflection.

Although farms do not traditionally have many women, I found this one to be female-friendly. The farm manager and co-owner were women. Additionally, my first impression of the robot came with the origin story of the robot's name, "Amiga." When the farmworkers were asked what they would like the robot's name to be, they named it the "Amiga," meaning a "female friend" in Spanish because the female farmer workers wanted more female co-workers. This also gave me an understanding of the values on the farm and made me more comfortable, as a woman, being there and designing.

Perhaps the most significant aspect of working in situ was having immediate access to the roboticists and farmers, which allowed us to negotiate designs on the spot. I have verbatim transcripts of these discussions because they were captured as part of my recordings of interactions people were having with the robot. In contrast, the conversations I had with myself had to be noted after the fact. When I had quiet moments, I sometimes recorded myself talking, but that could have been socially awkward if people had caught me doing it.

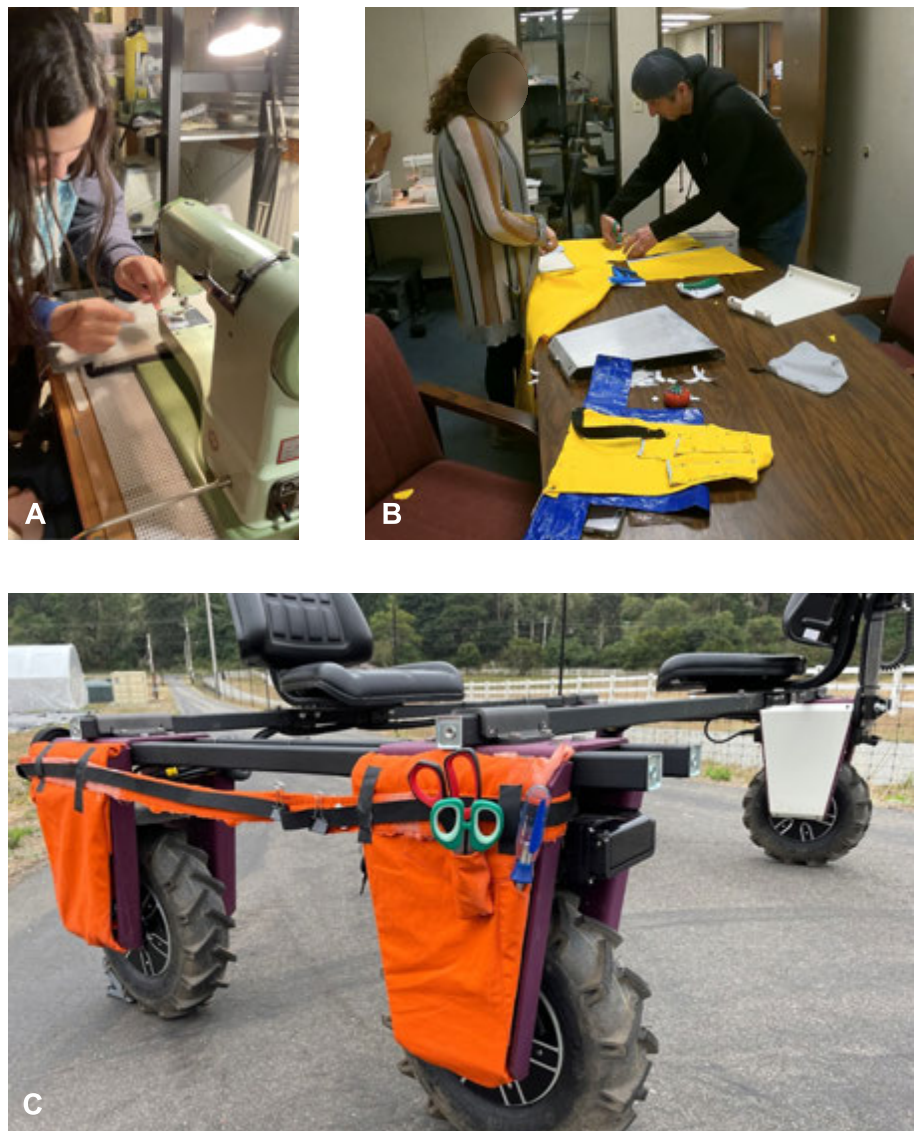

Fig. 13. Researcher learning from local practitioners, such as a:
(A) sewing practitioner,
(B) robotics designer. The bottom picture shows,
(C) the first prototype discussed with a sewing practitioner

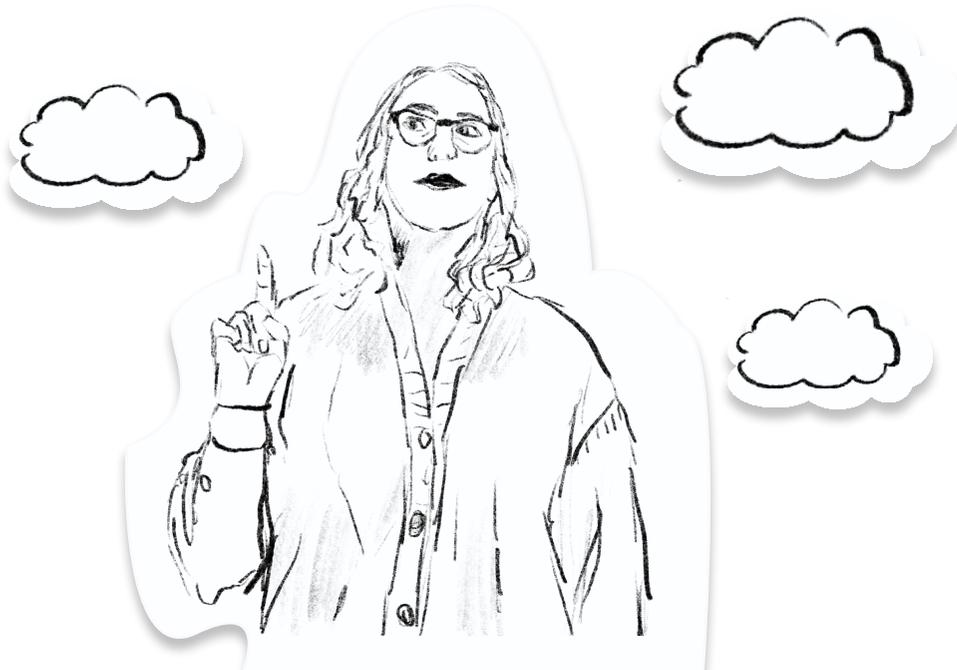

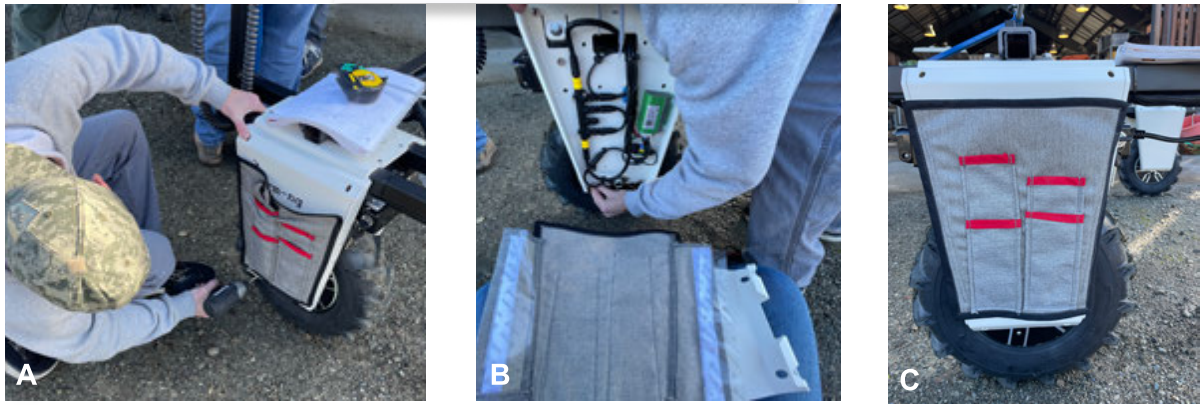

**Fig. 14. Installation of the latest version of the utility belt.**
**(A) Robot engineer puts on plates after the utility belt is strapped on.**
**(B) Engineer prepares the robot for the utility belt while the researcher sits on the ground with the utility utility belt on her lap**
**(C) Utility belt on the robot at an educational farm**

### 5.3 Other reflections on method

***Documentation***. This pictorial is possible because we planned to document everything from the start thoroughly. Nevertheless, we later found moments that were not captured as exhaustively as we had hoped. It would have been nice to have a team of photographers or a special robot just documenting my process. This illuminates how valuable the process of treating reflection-in-action can be not only for research but also for design. The documentation process has given us tools to share with other design researchers to refine the craft.

***Clothing is contextual.*** Designing clothes for the tractor robot helped us to reflect on what is also happening when we design clothes for people. Our robot clothing was intentionally crafted to afford its mechanical and social function. However, a lot of what we did was take in information about the context, the environment, the people, and the activities and try to manifest these discoveries into our decisions for the robot's utility belt. Fashion is sometimes looked down upon as mere decoration, but in fact, similarly reflects what designers notice about culture and the broader context. Perhaps robot clothes have the potential to expand and disrupt our norms within fashion for people.

## 6 CONCLUSION

While our object of design—the robot utility belt— is novel, we believe that the practice of in situ design is quite common. We note, however, that the documentation and reflection of in situ design are not as prevalent in the research-through-design literature as they could be. By showcasing our processes and reflections, we aim to inspire other design researchers to capture and share their own experiences of designing in the field.